\documentclass[conference]{IEEEtran}
\usepackage{times}

\usepackage[numbers]{natbib}
\usepackage{multicol}
\usepackage[bookmarks=true]{hyperref}
\usepackage{graphicx,subcaption} 
\usepackage{amssymb}
\usepackage[mathscr]{euscript}
\usepackage{algorithm2e}
\RestyleAlgo{ruled}
\usepackage{amsthm}
\theoremstyle{definition}
\newtheorem{definition}{Definition}[section]
\usepackage{xcolor}
\let\labelindent\relax
\usepackage{stfloats,microtype,balance,enumitem}
\usepackage[some,bottom]{background}

\usepackage{xspace}

\begin{document}

\title{
%Composing Independent Options Requires Retraining: Initial Results
%~\\
%Adapting Pre-trained Options : Initial Results
%~\\
%Composing Option Sequences Using Retraining: Initial Results
%~\\
Composing Option Sequences by Adaptation: Initial Results
%~\\
%Option Reuse: Initial Results
%~\\
%Retread Options: Going further with what you already have
%~\\
}

% You will get a Paper-ID when submitting a pdf file to the conference system
% \author{Author Names Omitted for Anonymous Review. Paper-ID 18}

% \author{\authorblockN{Charles A. Meehan}
% \authorblockA{U.S. Naval Research Laboratory\\
% Washington, D.C. \\
% United States\\
% charles.a.meehan2.civ@us.navy.mil}
% \and
% \authorblockN{Paul Rademacher}
% \authorblockA{U.S. Naval Research Laboratory\\
% Washington, D.C.\\
% United States\\
% paul.g.rademacher.civ@us.navy.mil}
% \and
% \authorblockN{Mark Roberts}
% \authorblockA{U.S. Naval Research Laboratory\\
% Washington, D.C.\\
% United States\\
% mark.c.roberts20.civ@us.navy.mil}
% \and
% \authorblockN{Laura M. Hiatt}
% \authorblockA{U.S. Naval Research Laboratory\\
% Washington, D.C.\\
% United States\\
% laura.m.hiatt.civ@us.navy.mil}}

% avoiding spaces at the end of the author lines is not a problem with
% conference papers because we don't use \thanks or \IEEEmembership

\author{\authorblockN{Charles A. Meehan, Paul Rademacher, Mark Roberts, and Laura M. Hiatt}
\authorblockA{U.S. Naval Research Laboratory\\
Washington, D.C.\\
United States\\ Emails: charles.a.meehan2.civ@us.navy.mil, paul.g.rademacher.civ@us.navy.mil, \\ mark.c.roberts20.civ@us.navy.mil, and laura.m.hiatt.civ@us.navy.mil}
}

\maketitle

\SetBgContents{\begin{small}\textbf{DISTRIBUTION STATEMENT A.} Approved for public release. Distribution unlimited.\end{small}}
\SetBgScale{1}
\SetBgOpacity{1}
\SetBgVshift{1cm}
\SetBgColor{black}
\BgThispage

\begin{abstract}
Robot manipulation in real-world settings often requires adapting the robot's behavior to the current situation, such as by changing the sequences in which policies execute to achieve the desired task.
Problematically, however, we show that composing a novel sequence of five deep RL options to perform a pick-and-place task is unlikely to successfully complete, even if their initiation and termination conditions align. We propose a framework to determine whether sequences will succeed {\em a priori}, and examine three approaches that adapt options to sequence successfully if they will not. Crucially, our adaptation methods consider the actual subset of points that the option is trained from or where it ends: (1) trains the second option to start where the first ends; (2)  trains the first option to reach the centroid of where the second starts; and (3) trains the first option to reach the median of where the second starts. Our results show that our framework and adaptation methods have promise in adapting options to work in novel sequences.  
\end{abstract}

\IEEEpeerreviewmaketitle

\section{Introduction}

Robot manipulation in real-world settings often requires adapting the robot's behavior to the current situation, such as by changing the sequences in which policies execute to achieve the desired task. This can potentially require learning new policies that sequence together in the new order.
Training policies for robot manipulation from scratch, however, can be very expensive, threatening a robot's adaptability. It also disregards the existing (if not perfect) functionality of the existing policies that work together to accomplish the task (e.g., policies from an options framework \cite{SUTTON1999181, bacon2016optioncritic, composable_parameterized_skills, symbolicRepParamSkills, fromSkillsToSymbols, skill_discovery, bagaria_21, Patra_Cavolowsky_Kulaksizoglu_Li_Hiatt_Roberts_Nau_2022}), even though they likely won't work in new sequences as-is \cite{logical_options_framework, kumar2024practice}.
In this paper, we aim to address this issue by proposing a framework for determining whether options can execute in new sequences, and proposing methods for re-using existing options by efficiently {\em retraining} them to perform in novel sequences.

\begin{figure}[t]
\includegraphics[width=\columnwidth]{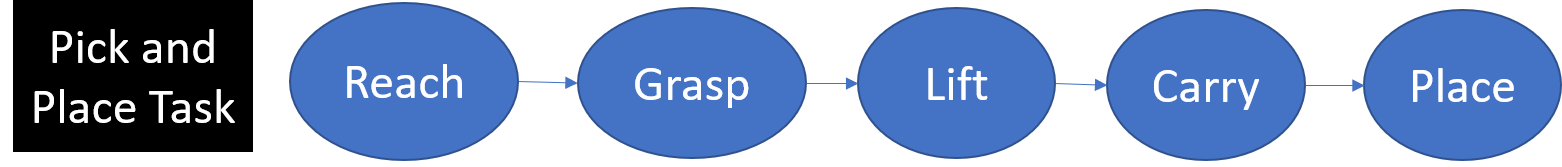}
\centering
\caption{Sequence of options for a pick-and-place task.}
\label{fig:pickandplace}
\end{figure}

To highlight the difficulty of ordering options in novel sequences, we train five options that, in theory, sequence together to perform a pick-and-place task (Figure~\ref{fig:pickandplace}).  Accordingly, each pair in this sequence has overlapping termination (TERM) and initiation (INIT) conditions. We capture the idea that they are being executed in a novel sequence by training them completely independently. Despite the connections between their termination and initiation conditions, these five options catastrophically fail to correctly execute the pick-and-place task; even pairs of them are unable to correctly execute.

In this paper, we provide evidence that the key reason for this failure has to do with a mismatch between where the first option in a sequence {\em actually ends after execution}, and where the next option is {\em actually trained to begin execution}. This suggests that the sets of where the actual execution of options begin and end are meaningful subsets of the formal initiation and termination conditions, and thus need to be explicitly considered when ordering options in novel sequences. We provide a framework for distinguishing between these conditions and sets in Section~\ref{sec: composablesec}. We then introduce three methods to adapt  options to improve their performance when executing in sequence in Section~\ref{sec:adaptationsection}. Next, we evaluate the adaptation methods in terms of execution success and training samples in Section~\ref{sec:evaluation}. Our results show that our framework and adaptation methods have much promise in adapting options to work in novel sequences, and suggest several directions for future exploration of this topic.

\section{Problem Setup} \label{problemsetup}

%\subsection{Background: The Options Framework}

\newcommand{\mdpstates}{\ensuremath{\mathcal{S}}\xspace}
\newcommand{\mdpactions}{\ensuremath{\mathcal{A}}\xspace}
\newcommand{\mdpreward}{\ensuremath{\mathcal{R}}\xspace}
\newcommand{\mdptransition}{\ensuremath{\mathcal{P}}\xspace}

\newcommand{\policy}{\ensuremath{\pi}\xspace}
\newcommand{\initcond}{\ensuremath{INIT}\xspace}
\newcommand{\termcond}{\ensuremath{TERM}\xspace}
\newcommand{\optionsingle}{\ensuremath{o}\xspace}
\newcommand{\option}[1]{\ensuremath{o_{#1}}\xspace}
\newcommand{\optioni}{\option{i}}
\newcommand{\optionj}{\option{j}}
\newcommand{\optionk}{\option{k}}
\newcommand{\policyi}{\ensuremath{\policy_i}\xspace}
\newcommand{\policyj}{\ensuremath{\policy_j}\xspace}
\newcommand{\optionreward}{\ensuremath{R}\xspace}
\newcommand{\st}{\ensuremath{\mid}}

\newcommand{\initsetsingle}{\ensuremath{I}\xspace}
\newcommand{\initset}[1]{\ensuremath{\initsetsingle_{#1}}\xspace}
\newcommand{\termsetsingle}{\ensuremath{T}\xspace}
\newcommand{\termset}[1]{\ensuremath{\termsetsingle_{#1}}\xspace}

\newcommand{\resultsetsingle}{\ensuremath{R}}
\newcommand{\resultset}[1]{\ensuremath{\resultsetsingle_{#1}}}
\newcommand{\originsetsingle}{\ensuremath{O}}
\newcommand{\originset}[1]{\ensuremath{\originsetsingle_{#1}}}

The standard set up for the reinforcement learning problem is as a Markov Decision Process or MDP \cite{SUTTON1999181}. A MDP model is made up of the following:
\begin{itemize}
    \item \mdpstates is a set of states.
    \item \mdpactions is a set of actions. 
    \item $\mdptransition_a(s,s') = \mdptransition_r(s_{t+1}\!=\!s'|s_t\!=\!s,a_t\!=\!a)$ is the probability that action a in state s at time t will lead to state s' at time t+1.
    \item $\mdpreward(s, a ,s')$ is the reward received after transitioning from state s to s' after taking action a.
\end{itemize}
The solution to an MDP is a policy $\policy : \mdpstates \rightarrow \mdpactions$ that maps states to actions. A temporally extended action, an option, is a sub-policy that starts executing when an initiation condition, $\initcond : \mdpstates \rightarrow \{true, false\}$ becomes true and stops executing when a termination condition $\termcond : \mdpstates \rightarrow \{true, false\}$ is true.   
An option is a tuple $\optionsingle = (\policy, \initcond, \termcond, \optionreward)$, where \optionreward is a reward function that may respect the overall reward \mdpreward.
We will often reference the states resulting from the application of \initcond as \initsetsingle and the states resulting from the application of the termination condition \termcond as \termsetsingle.  For a particular option  \optioni, these sets are as follows: \\ 
\phantom{xxxxx}  $\initset{i}  = \{ s \in \mdpstates \st \initcond(s) = true\}$,  and \\
\phantom{xxxxx} $\termset{i} = \{ s \in \mdpstates \st \termcond(s) = true\}$.

%\subsection{Example problem: One-armed Manipulation}

\begin{figure}[t]
\includegraphics[width=0.8\columnwidth]{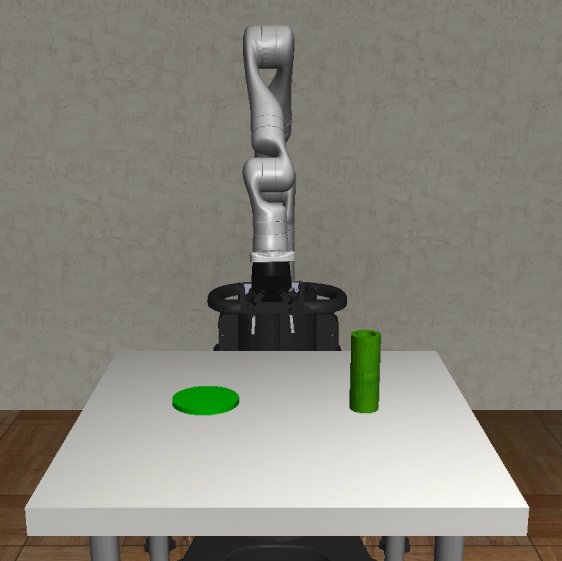}
\centering
\caption{Kinova Gen3 robotic arm in a simplified pick-and-place environment built in robosuite.}
\label{fig:kinova3sim}
\end{figure}

Our main objective in this study is to examine how well a connected sequence of pre-trained options completes a task.  For this study, we connect a sequence of five options\footnote{These options are actually implemented as goal skills, which are options extended with the addition of maintenance conditions that truncate a run if violated \cite{Patra_Cavolowsky_Kulaksizoglu_Li_Hiatt_Roberts_Nau_2022}.  Our implementation has maintenance conditions equal to the initiaton set of the skill, so we simplify the discussion in this paper to just the initiation and termination sets for this work.  Future work will extend this model to the full specification of goals skills.}. 
for a pick-and-place task using a single arm.  Figure~\ref{fig:pickandplace} shows the sequence of five options we manually created:  $Reach(item)$,  $Grasp(item)$, $Lift(item, target)$, $Carry(item, target)$,  and $Place(item, target)$.
For each option, we defined \initcond and \termcond such that for each adjacent pair of options $(\optioni, \optionj)$ in this sequence the termination set of the first option is equal to the initiation set of the second option; that is, $\termset{i} = \initset{j}$. Each option was trained independently in a simulated pick-and-place environment built in robosuite, a simulation framework powered by MuJoCo physics engine for robotic learning \cite{robosuite2020},  as shown in Figure~\ref{fig:kinova3sim}. Further explanation of training these five options will be discussed in Section \ref{sec:evaluation}.

\section{Connected and Composable Options} \label{sec: composablesec}

Consider two options \optioni and \optionj where  \optionj executes from the point where \optioni finishes.  
We say two options are \emph{connected} when the set of states from $\termcond(\optioni)$ is a subset\footnote{We focus on a strict subsets between sets for simplicity.  More generally, we could measure overlap with a metric such as Jaccard distance.} of the $\initcond(\optionj)$.  For the sets of these conditions, this is equivalent to $\termset{i} \subseteq \initset{j}$. Although these options may be connected, they might fail to  execute together in practice, which we demonstrate in our evaluation. The reason has to do with the fact that these options are independently trained.

\begin{figure}[t]
\includegraphics[width=\columnwidth]{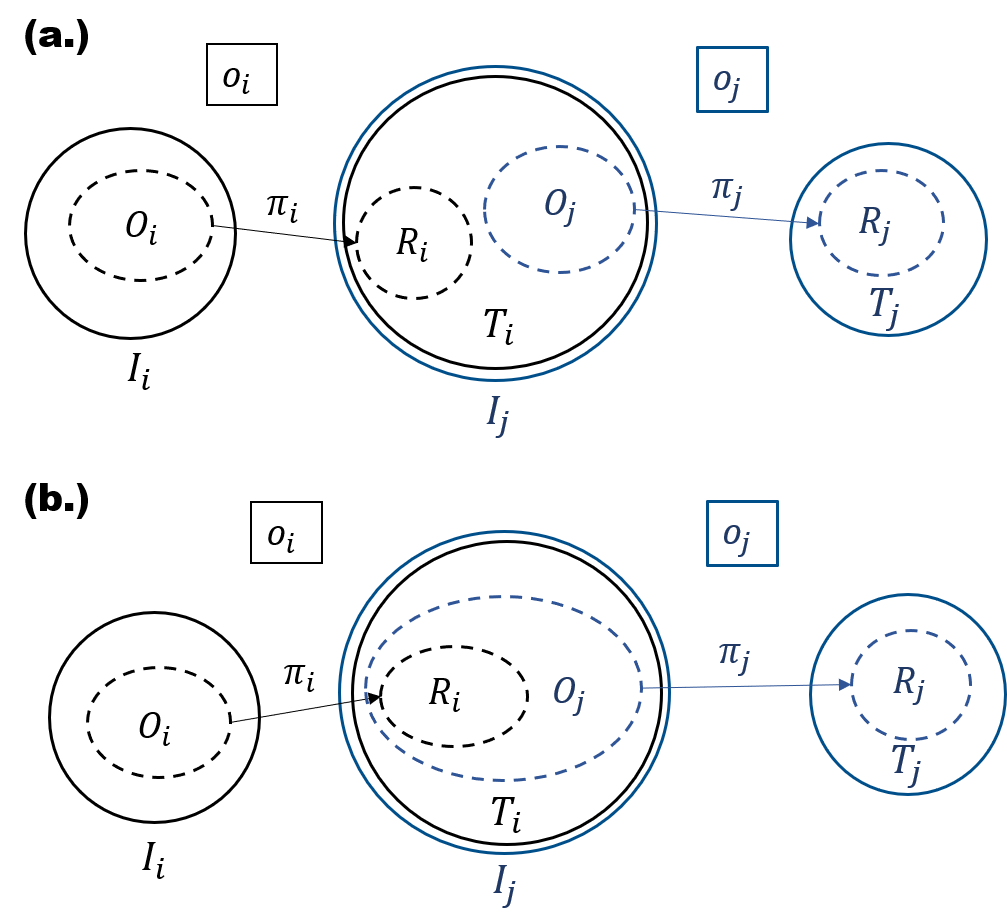}
\centering
\caption{Conceptual illustration of connected and composable options, $o_i$ and $o_j$. (a.) Connected options. (b.) Composable options.}
\label{fig:connectedcomposable}
\end{figure}

To see why, consider Figure~\ref{fig:connectedcomposable} (a).  
Let us call the result of running \policyi, the policy of \optioni, the result set \resultset{i}\footnote{The result set is a more constrained version of the effect set from \cite{fromSkillsToSymbols}. The option's policy can only initiate from states within the origin set not from any state in the agent's state space as assumed with the effect set.}.
Further, let us call the set of states used to train \optionj the origin set \originset{j}.
Even though $\termset{i} \subseteq \initset{j}$, the result set \resultset{i} does not overlap with  \originset{j}, resulting in the two options being misaligned even though they have connecting conditions.

To align the options, we need to adapt the options to match Figure~\ref{fig:connectedcomposable} (b), so that \optioni ends in a place that overlaps with the set of states that \optionj was trained to start from.  That is, train the options so that $\resultset{i} \subseteq \originset{j}$. Considering the origin and result sets, in contrast to the initiation and termination conditions, leads to our definition of composable options.

\begin{definition}[Composable Options] \label{Composablegoalskills}
Options, $o_i$ and $o_j$, are composable if $\resultset{i} \subseteq \originset{j}$.
\end{definition}

Figures \ref{fig:connectedcomposable}(a.) and \ref{fig:connectedcomposable}(b.) are illustrations of connected and composable options. We hypothesize that sequences of composable options will execute successfully more often, where success is measured as completing the final task. We test this hypothesis by measuring the performance success of the pick-and-place task by executing pairs and sequences of 3, 4, and 5 connected options. The method that we use to evaluate the performance success will be discussed in Section \ref{sec:evaluation}.

\section{Adapting Options}
\label{sec:adaptationsection}

\begin{figure}[t]
\includegraphics[width=\columnwidth]{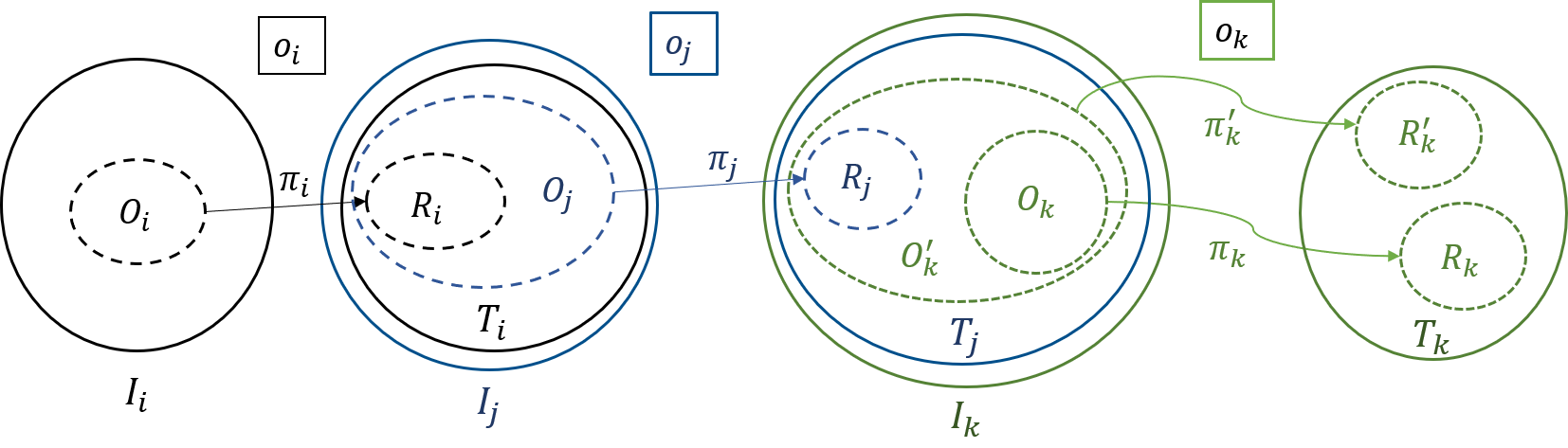}
\centering
\caption{Diagram of the adaptation of option, \optionk, using the Origin Method.}
\label{fig:incompatibleend}
\end{figure}

Connected options do not guarantee successful execution. We believe that this is due to the origin and results sets not overlapping.
Consider Figure \ref{fig:incompatibleend}, which shows three options \optioni, \optionj, and \optionk, where \optionj and \optionk are connected but not composable (i.e., $\resultset{j} \cap \originset{k} = \emptyset$).  
There are two main ways we can make them composable: adapt \originset{k} or adapt \resultset{j}.  In the following three subsections, we outline approaches that do one of these adaptations.

\subsection{Origin Method: Origin}\label{sec:dependentmethod}

Using Figure \ref{fig:incompatibleend} to visualize the adaptation of an option, the Origin Method (Origin for short) expands the set \originset{k} to be a new set $\originset{k}'$ so that $\resultset{j} \subseteq \originset{k}'$ thereby satisfying Definition \ref{Composablegoalskills}. This expansion of an origin set can be accomplished by starting a new training session where the agent starts in \originset{j} and follows a fixed policy, \policyj, which terminates in a state in the set, \resultset{j}. The agent then starts from this state and attempts to learn a policy that will terminate in the set \termset{k}. This training will continue until the policy has converged which results in a new policy, ${\pi}'_{k}$, which can start from any state in $\resultset{j} \subseteq \originset{k}'$ and terminate in a new set $\resultset{k}' \subseteq \termset{k}$. This method is called the origin method because we are expanding the origin set of the second option so it overlaps with the result set of the previous method.

\begin{figure}[t]
\includegraphics[width=\columnwidth]{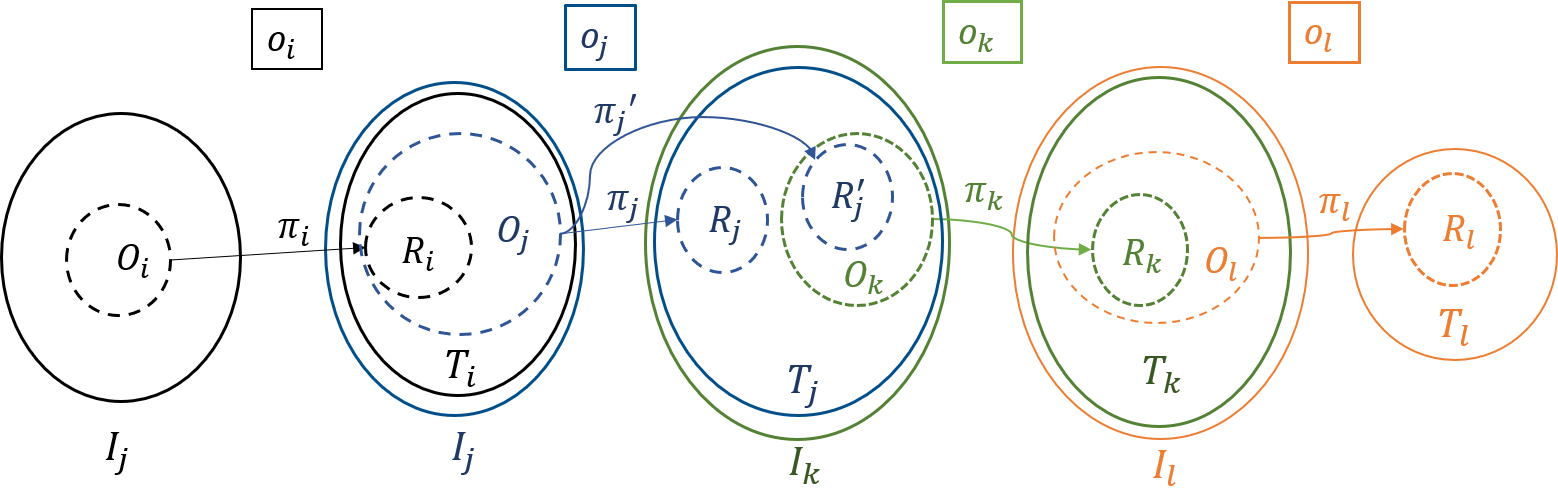}
\centering
\caption{Diagram of the adaptation of option, \optionj, using the Result Methods.}
\label{fig:resultMethodAdapt}
\end{figure}

\subsection{Result Methods: RM-Centroid and RM-Density}

Figure \ref{fig:resultMethodAdapt} demonstrates the adaptation of an option, \optionj, using the Result Methods. The objective of both Result Methods is to move the result set of an option to be a subset of the origin set of the succeeding option in the sequence. In Figure \ref{fig:resultMethodAdapt}, the result set, \resultset{j}, is moved to be a subset of origin set, \originset{k}. We tested moving the result set towards the centroid of the succeeding origin set (RM-Centroid) and moving the result set towards the most dense area of the succeeding origin set (RM-Density).

Using the hypothetical sequence pictured Figure \ref{fig:resultMethodAdapt}, RM-Centroid first finds the centroid of the origin set, \originset{k}, by collecting samples of the start states used in training policy, $\pi_k$. The sample closest to the centroid is then set as the goal for the new policy, $\pi_j'$, to move the result set, \resultset{j}, towards. After retraining, the new result set, $\resultset{j}'$ should be a subset of \originset{k}. RM-Density used the same sampled origin sets that RM-Centroid used and the only difference is that the sample chosen as the goal is the sample in the most dense region of the origin set. The details of the collection of samples and how the options were adapted by these methods are discussed in Section \ref{sec:evaluation}.

\section{Evaluation} \label{sec:evaluation}

Before discussing our process for evaluating the performance of the connected options, we review our setup for training the five independent options for the pick-and-place task and the sampling and training processes used for the Result Methods.

\subsection{Independent Training Setup}

Figure~\ref{fig:kinova3sim} shows our implementation of the task using robosuite \cite{robosuite2020}. We used a 7DOF Kinova Gen3 arm \cite{kinovagen3}, and we trained the options using SAC algorithm \cite{haarnoja2018sac} in Stable Baselines3 (SB3) \cite{stable-baselines3}. We created the necessary training scripts to connect the robosuite environments to SAC algorithm from SB3. The action space used during training the robotic arm was equal to the task space ($\mathscr{T}$) of the end effector which is the space of all possible end-effector poses, $\mathscr{T} \subset SE(3)$. The observation space used during training was equal to the proprioception information from the robot arm such as the arm joint positions and velocities, end effector pose, gripper joint positions and velocities. The object information that was included in the observation space was the x, y, z positions of the table, green target, and green cup. Also, the context of the cup being full or empty was included as -1 for empty, 0 for none, and 1 for full. 

Model-free algorithms like SAC are known to struggle with policy convergence when the agent is only given sparse rewards \cite{haarnoja2018composable} such as only receiving a reward for completing a goal in a high dimensional action or observation space. Since the pick-and-place task has both a high dimensional action and observation space, reward shaping functions were used to encourage the five options to reduce the distance between the end effector and an object or target. The reward shaping function used for these options which is a modified version of the reward shaping function from \cite{robosuite2020} is

\begin{equation}\label{eq:rewardshapingfunction}
    reward = -1 * \tanh(dist)^2
\end{equation}

\noindent where dist is equal to the Euclidean distance between the end effector and an object or a target depending on the option. A reward of 1000 is given to the agent when the termination conditions of an option are satisfied. Each option is trained independently until convergence. Convergence for these policies was when a max reward threshold was surpassed during evaluation of the trained policy. The best model (learned optimal policy) was saved for each option.

\subsection{Adaptation Details of RM-Centroid and RM-Density}

\begin{figure}[t]
\includegraphics[width=\columnwidth]{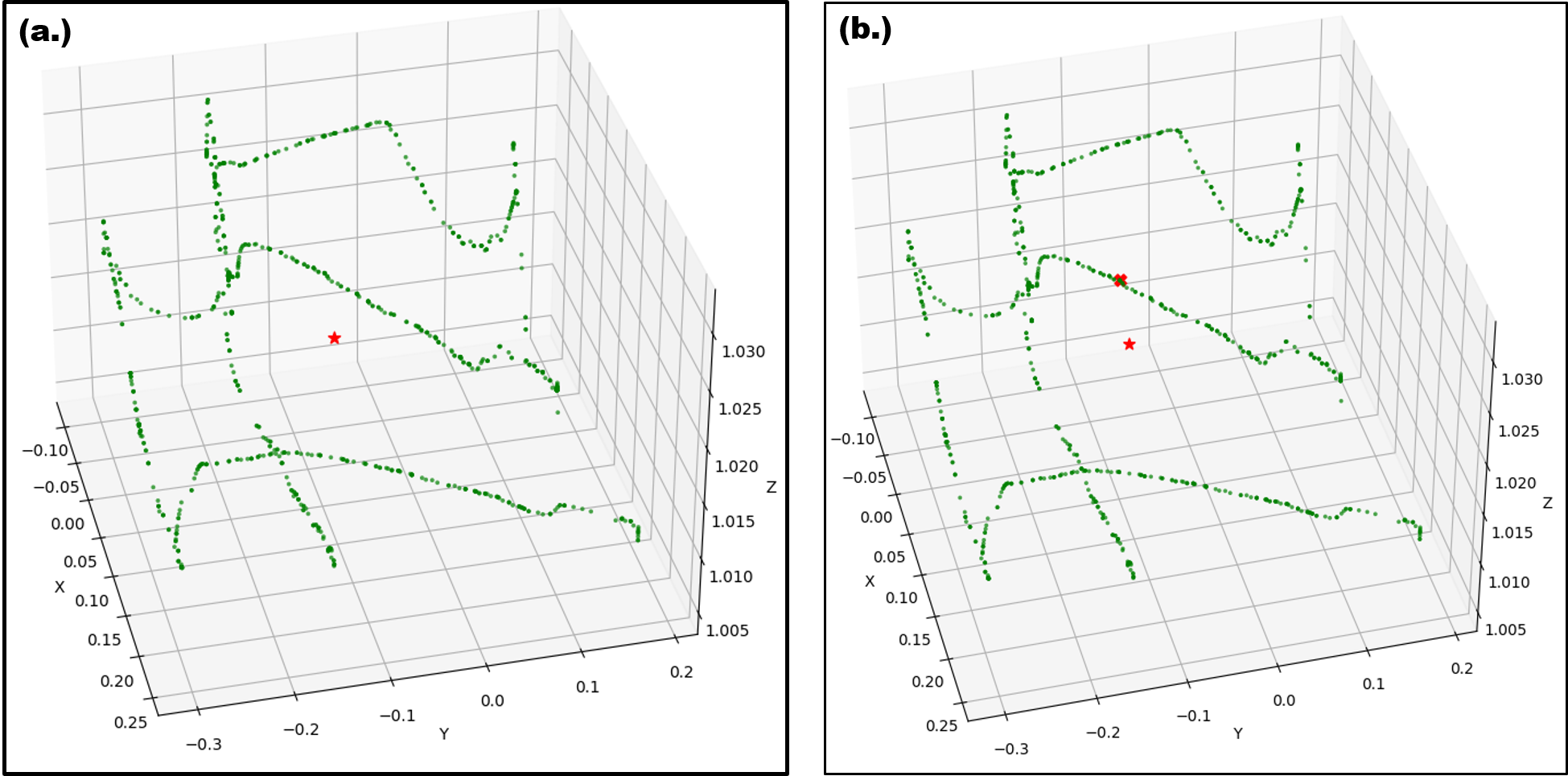}
\caption{(a.) 1000 samples recorded from the carry option origin set. (b.) Carry option origin set with centroid marked as the red star and the closest sample marked with a red plus in red.}
\centering
\label{fig:carryoriginsetandcentroid}
\end{figure}

In order to use the adaptation methods of RM-Centroid and RM-Density, it is necessary to sample the origin sets of all five independently trained options. An example of 1000 samples collected for the origin set of the carry option is shown in Figure \ref{fig:carryoriginsetandcentroid}(a.). Each recorded sample is the position and orientation of the end effector, but only the positional data is visualized in Figures \ref{fig:carryoriginsetandcentroid}(a.) and (b.). 

For RM-Centroid, the centroid of the sampled set is found by taking the average position of along the three positional axes. The centroid for the carry option is the red star in Figures \ref{fig:carryoriginsetandcentroid}(a.) and (b.). The next step of RM-Centroid is to find the sample that is closest in Euclidean distance to the centroid of the sampled origin set. The sample that is closest to the centroid is circled in red in Figure \ref{fig:carryoriginsetandcentroid}(b.). A new reward shaping function is used to motivate the robotic arm to move its end effector to end in a position and orientation that is very close to the position and orientation of this sample. The new reward shaping function, Function \ref{eq:rewardfunction}, is

\begin{equation}\label{eq:rewardfunction}
    reward = -10 * \tanh(ee_{pose\_dist\_centroid})^2
\end{equation}

\noindent $ee_{pose\_dist\_centroid}$ is the Euclidean distance of the end effector to the centroid sample position plus the orientation difference as shown in Function \ref{eq:posedistancecentroid}.

\begin{equation}\label{eq:posedistancecentroid}
    ee_{pose\_dist\_centroid} = \|position_{centroid} - position_{ee}\| + \rho
\end{equation}

\noindent $\rho$ is the orientation difference between two quaternions which is equal to zero when the quaternions are in the same direction and one when they are 180 degrees apart. The formula for $\rho$ is given by 

\begin{equation}\label{eq:rho}
    \rho = \omega_r * (1 - \|Quaternion_{ee} \cdot Quaternion_{centroid}\|)
\end{equation}

\noindent which is borrowed from \citet{orientationdifferencepaper}. $\rho$ is on the range $[0, \omega_r]$ where $\omega_r$ is typically chosen to be equal to one. 

\begin{figure}[t]
\includegraphics[width=0.85\columnwidth]{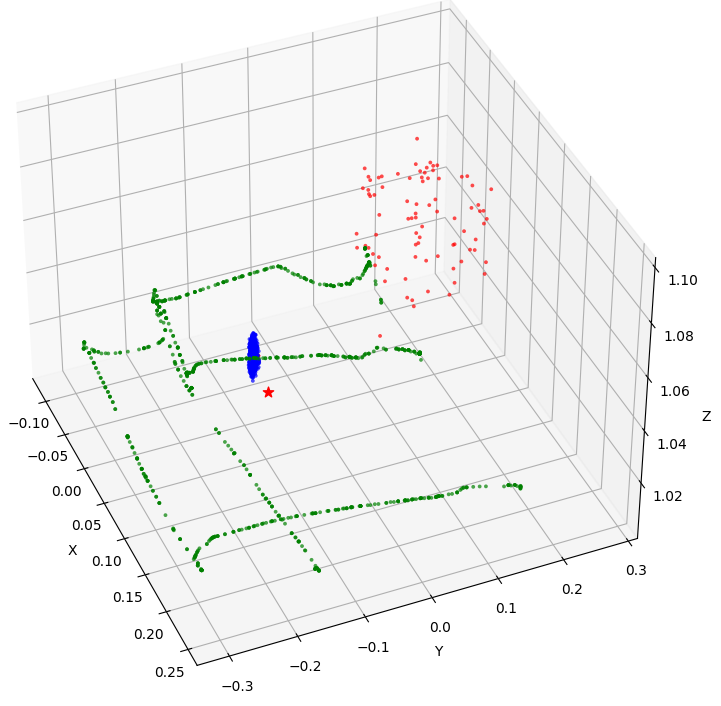}
\centering
\caption{Samples taken from the independently trained lift option in red, lift option adapted by the RM-Centroid method in blue, and the carry option origin set in green.}
\label{fig:liftAdaptedDep}
\end{figure}

A positive reward of 1000 is given to the agent when the minimum distance is achieved ($ee_{pose\_dist\_centroid} \leq 0.01$), and the termination conditions of the option are satisfied. An example of adapting the result set of the lift option to be within 0.01 distance to the selected sample from the carry option origin set is shown in Figure \ref{fig:liftAdaptedDep}. The red samples are samples of the result set of the independently trained lift option. The blue set of samples is the result set of the adapted, origin lift option, and the green samples are the carry option origin set with the centroid denoted by the red star. It is clear in the figure that RM-Centroid is able to move the result set of the independently trained option to be close to the selected sample that was closest to the centroid of the origin set.

\begin{figure}[t]
\includegraphics[width=0.85\columnwidth]{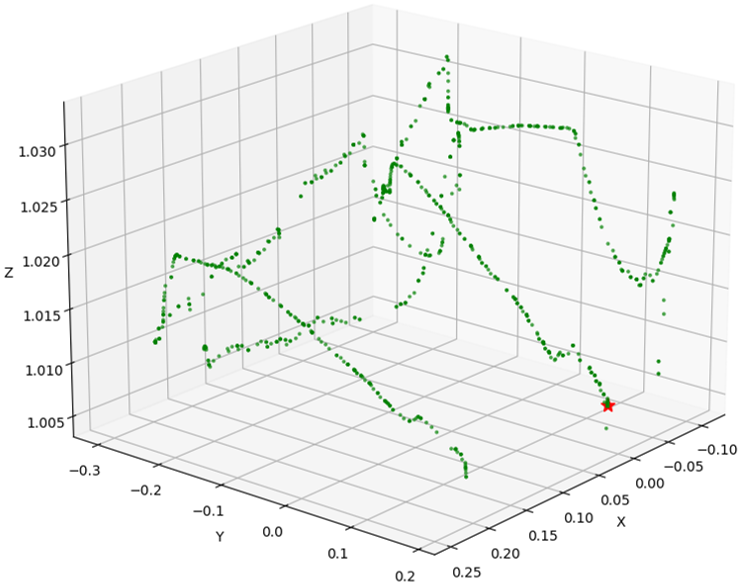}
\centering
\caption{Carry option origin set with a red star over the dense sample selected by the RM-Density method}
\label{fig:densebin}
\end{figure}

For RM-Density, it is necessary to find the sample in the most dense region of the origin set. The distance function as defined in Equation \ref{eq:posedistancecentroid} was used to calculate the distance between every sample of the set of samples. The sample with the highest number of close neighbor samples (samples that had the smallest distance to that sample) was chosen as the sample in the most dense region of the origin set. An example of this is shown in Figure \ref{fig:densebin} for the carry option origin set. The method of training was the same as done for RM-Centroid with this sample's position and orientation selected as the goal to achieve.

\subsection{Measuring Performance of Connected Options}

Once the independent policies for each of the five options had converged, we measured the number of times the last option in a sequence of connected options successfully terminated out of the total number of times the connected options were executed which gave us a measure of performance. Sequences of 2, 3, 4, and 5 connected options were evaluated. Each sequence was executed for 100 sets of 10 episodes. An episode would terminate only when the termination conditions of the final option in the sequence were satisfied or if an action was taken that violated a environmental condition. We set three environmental conditions that applied to all options which were:
\begin{itemize}
    \item The robot arm could not collide with itself or any part of the environment.
    \item The orientation of the cup could not be horizontal (cup could not be knocked over).
    \item The position of the cup could not more than 0.4m away from the edges of the table.
\end{itemize}

The blue bars in Figure \ref{fig:compatiblilityplot} show the rate of success for each of the sequences of connected independently trained options. As hypothesized, connected options do not guarantee a high execution success rate when connecting independently trained options. Only the connected pair of grasp and lift options has a performance rate better than 50\%. We believe that this low performance rate is an indicator that the origin set and result set of each connected option are not fully or are only partially overlapping. Therefore, Definition \ref{Composablegoalskills} is not satisfied. 

\begin{figure}[t]
\includegraphics[width=\columnwidth]{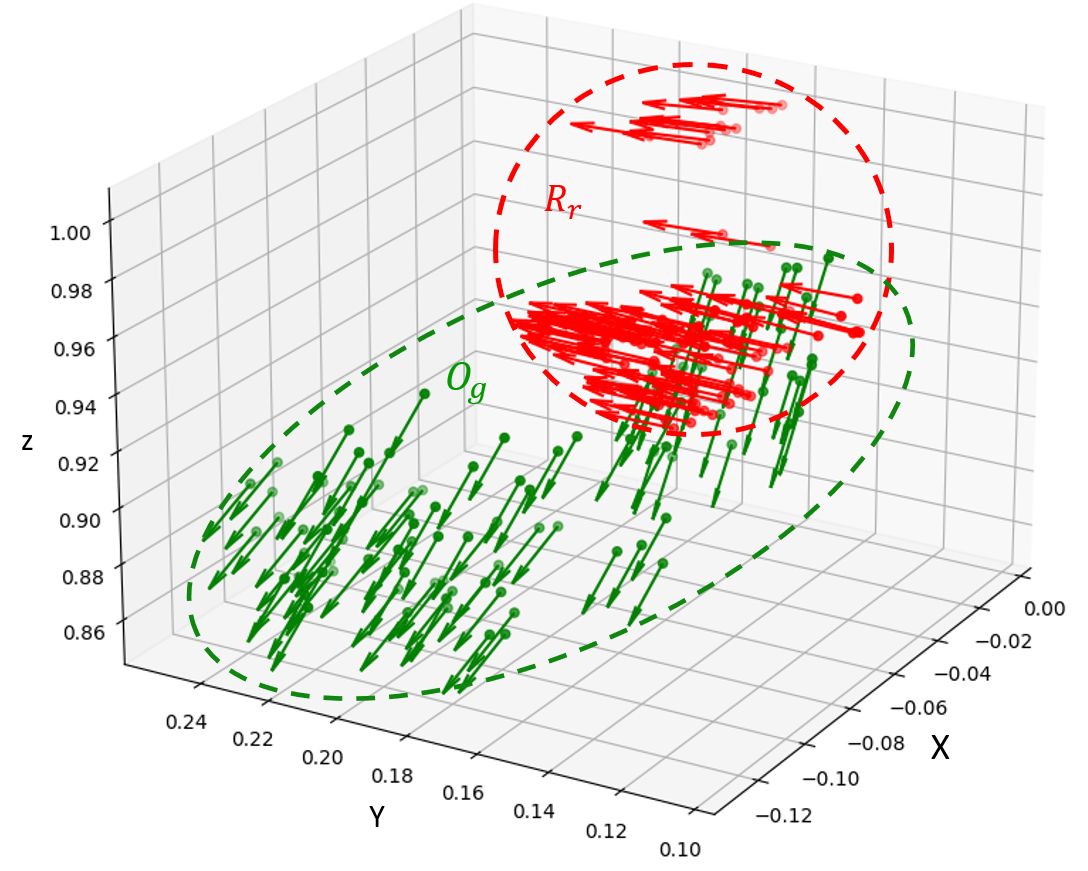}
\centering
\caption{Sampled reach option result set in red and sampled grasp origin set in green.} 
\label{fig:sampledSets}
\end{figure}

This is supported by an example visualization in Figure \ref{fig:sampledSets} of the origin set and result set for the reach and grasp options which had a zero performance rate. This plot shows the sampled result for reach and origin set for grasp. The samples are made up of the the x, y, and z position of the end effector. The orientation of the end effector is represented by the arrows on the plot. As you can see these sets appear to not fully satisfy Definition \ref{Composablegoalskills}. To test that modifying the origin set or the result set in order to satisfy Definition \ref{Composablegoalskills} will improve performance of connected options, we adapted the options using the Origin method, RM-Centroid, and the RM-Density. We discuss the measured performance results of these methods in Section \ref{sec:results}. 

% Add comparison plots here

\subsection{Results}
\label{sec:results}
\begin{figure*}[t]
\includegraphics[width=0.8\textwidth]{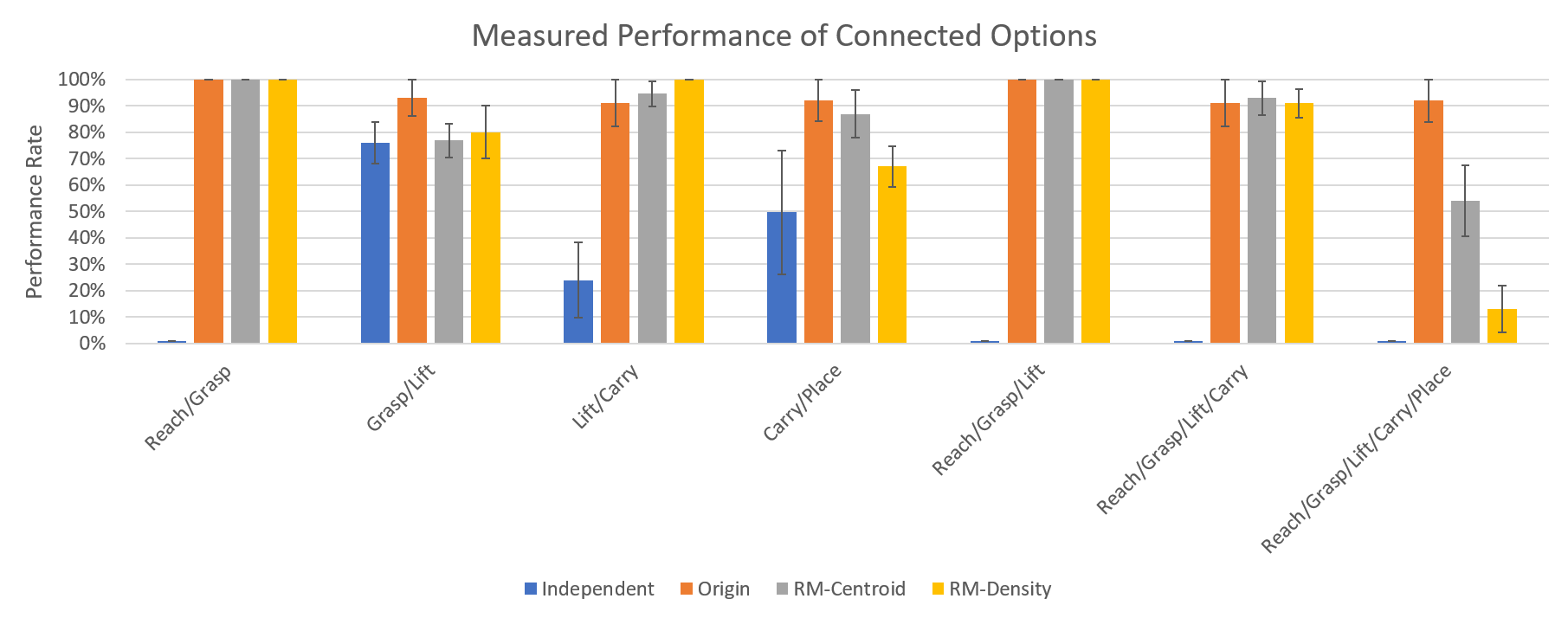}
\centering
\caption{Performance results for connected independent options and adapted options.}
\label{fig:compatiblilityplot}
\end{figure*}

Figure \ref{fig:compatiblilityplot} shows the success rate for the  independently trained options (blue) and the options adapted by the origin method (orange), the RM-Centroid method (gray), and RM-Density (gold).     
As stated previously, the connected independently trained options do not perform well: it does not even reach 5\% success at any sequence greater than three. 
The origin method performs at or better than most of the other approaches.  For pairs of tasks the results are mixed as seen by the high performance rate of the result methods even being equal or higher than the origin method for almost all pairings except for the pairing of the carry and place options. But for the final five-skill sequence, the origin method is the best performing. The RM-Centroid and RM-Density approaches have mixed results for smaller sequences but do not scale to the full five-skill sequence.

\begin{figure*}[t]
\includegraphics[width=0.8\textwidth]{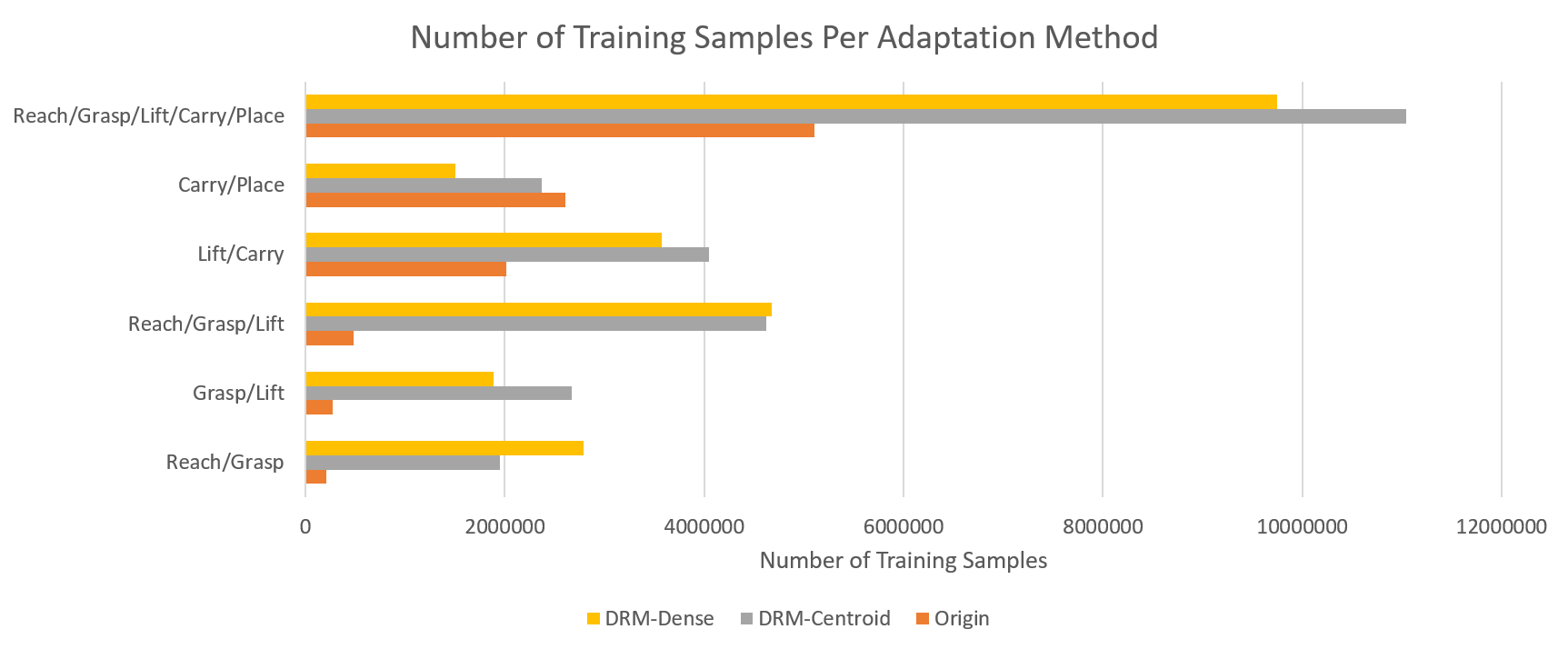}
\centering
\caption{Number of training samples needed till policy convergence given the three adaptation methods.}
\label{fig:trainingsamples}
\end{figure*}

We wanted to compare the sample complexity of the three adaptation approaches. Since a new policy and result set is created when expanding the origin set during the application of the Origin method, this method would need to be applied to all options in a sequence even to already composable pairings. This is not necessary when using the Result Methods for adaptation. Therefore, our original hypothesis was that RM-Centroid or RM-Density would result in a lower sample complexity. However, Figure~\ref{fig:trainingsamples} instead shows that the Origin Method usually resulted in the lowest sample complexity. We believe that this is due to the added complexity of training to a specific position and orientation when using the Result Methods. Although, we do believe that this gap will close once the approaches for the Result methods are enhanced and sequences of options are longer and more complex.

Overall, our results demonstrate that the Origin Method was best able to adapt the options and usually it does so with lower sample complexity.  This suggests that for now the origin method is the best method for adapting options to execute together even when those options are trained independently. 

\section{Related Work}

We have discussed the assumed successful execution of connected options in temporal abstraction methods that are based on the Options Framework \cite{SUTTON1999181}. This has led to methods that can discover options and chain them together to complete more difficult tasks \cite{skill_discovery, bagaria_21}. For two options, \optioni and \optionj, $\termset{i} \subseteq \initset{j}$ needs to be satisfied for composability. In the option chaining methods of \citet{bagaria_21}, options are trained to satisfy $\termset{i} = \initset{j}$ and are therefore composable but dependent on one another. One downside of the original option chaining work is not being able to provide any guarantees on the optimality of the learned chains of options. In more recent work, \citet{bagaria_robust_composable_skills, bagaria_23} have focused on ways to improve the optimality of the learned chains of options and make these chains more robust to changes in the initiation sets during training. 

While our work assumes a sequence of options is provided by a task planner instead of discovering these sequences, the initiation set classifiers \cite{bagaria_robust_composable_skills} or initiation value functions \cite{bagaria_23} used to determine if states are within the learned initiation sets could potentially be used in our adaptation work. If we can classify how far a state is from being within an origin set of an option, we could use this information when adapting a result set to be within an origin set of a connected option. In \cite{vats_recovery_learning}, recovery policies are learned to be able to recover to initiation sets from clusters of states that are deemed failures during execution of an option. \citet{vats_recovery_learning} learn multiple recovery policies from a failure cluster and towards the end of recovery policy training select the best learned recovery policy. This could be another way to improve our own adaptation methods. These and possible variations of these methods will be explored in future work.

\citet{abbatematteo2024composable} compose manipulation options through the use of a motion planner to connect two states by the means of a motion plan but could result in unwanted behavior in the hand-off between options. We chose in this work to adapt pretrained options to work together instead of adding an additional motion plan to connect the result and origin sets of trained options. 

There have been various methods for composing options through the use of learning a higher level option or controller that can select sequences of options that can complete tasks such as in \cite{bacon2016optioncritic, logical_options_framework}. However, these types of methods will require retraining of full sequences if approached with tasks that are outside of the training set. \citet{mendez2022composuite} created an compositional reinforcement learning suite of benchmarks to compare against but currently use a variant of learning higher level controllers that learn sequences of skills. In future work, we are looking into using these benchmarks as possible comparison to our composition of independent options when faced with new tasks.

Part of the motivation for using independently trained options comes with the release of robots like the Spot\textsuperscript{\textregistered} from Boston Dynamics \cite{spotrobot} in 2020. Consumers in warehouses, factories, and households can more readily purchase robots that arrive with a certain set of options. As pointed out in \cite{kumar2024practice}, consumers should be able to customize these robots to complete specific jobs which usually requires retraining of these previously trained options or the full replacement of these hard coded options. A lot of retraining or the worst case of full replacement of options could significantly delay the usefulness of these deployed robots. Instead of replacing pretrained options, \citet{kumar2024practice} focus on retraining options. However, they assume that any given sequence of options needed to complete a task are always composable. Therefore, the adaptation of the options is restricted by this assumption because if the option changes too much, then composability of the sequence of options is not guaranteed. We agree with the need to adapt independently trained options to complete new tasks, but we do not assume these options are composable. This makes the problem more difficult, but allows us more freedom to drastically adapt the behavior of the pretrained options when necessary.

\section{Conclusion} 
\label{sec:conclusion}

We showed that connected options -- i.e., options that have overlapping \initcond and \termcond -- fail to successfully complete a pick-and-place task. 
We defined composability of options in terms of two new sets for options called the origin and result sets. We hypothesized that the overlap of the origin and result sets of options was more important to the successful execution of sequences of options than the options just being connected. We created three methods (Origin Method, RM-Centroid, and RM-Density) to adapt the origin and result sets in order to satisfy the composability definition of options and test our hypothesis. We applied these three methods to our pick-and-place options and measured their rate of performance success. It was found that the Origin Method resulted in the highest rate of performance for the full pick-and-place task of all five connected options. 

We will continue to investigate the relationship between the overlap of the origin and results sets of connected options and the performance of them executing in sequence. If such a relationship is confirmed, it would not only increase the understanding of how to adapt options to better execute in sequence, but could also be used to {\em predict} the performance of connected options ahead of execution. 
We also plan to integrate these pre-trained and adapted options with a planning system (e.g.,  $GTPyhop^2$ \cite{nau2021gtpyhop}) to study more complex combinations of options.  For example, a planner might reuse the reach or grasp options from Figure~\ref{fig:pickandplace} in a plan that is used to open a door.  Such integration could help further test our adaptation methods in novel or longer sequences of options.

\section*{Acknowledgments}
The authors thank the Office of Naval Research and the U.S. Naval Research Laboratory for funding this research.
%% Use plainnat to work nicely with natbib. 

\bibliographystyle{plainnat}
\bibliography{references}

\onecolumn

\end{document}